\UseRawInputEncoding
\documentclass[11pt]{article}
\usepackage[a4paper]{geometry} 
\usepackage{clic2023} 
\usepackage{listings}
\usepackage{times} 

\usepackage{xurl} 

\usepackage[italian,english]{babel}

\usepackage{latexsym} 

\usepackage{xcolor}

\usepackage{graphicx}

\usepackage{algorithmic}

\usepackage{latexsym}

\usepackage{amsmath}

\usepackage[unboxed]{cwpuzzle}

\usepackage{booktabs}

\usepackage{xcolor}

\definecolor{codegreen}{rgb}{0,0.6,0}
\definecolor{codegray}{rgb}{0.5,0.5,0.5}
\definecolor{codepurple}{rgb}{0.58,0,0.82}
\definecolor{backcolour}{rgb}{0.95,0.95,0.92}

\lstdefinestyle{mystyle}{
  backgroundcolor=\color{backcolour}, commentstyle=\color{codegreen},
  keywordstyle=\color{magenta},
  numberstyle=\tiny\color{codegray},
  stringstyle=\color{codepurple},
  basicstyle=\ttfamily\footnotesize,
  breakatwhitespace=false,         
  breaklines=true,                 
  captionpos=b,                    
  keepspaces=true,                 
  numbers=false,                    
  numbersep=5pt,                  
  showspaces=false,                
  showstringspaces=false,
  showtabs=false,                  
  tabsize=2
}

\usepackage{setspace}
\lstdefinestyle{mystyle}{
  backgroundcolor=\color{backcolour},
  commentstyle=\color{codegreen},
  keywordstyle=\color{codepurple},
  numberstyle=\tiny\color{codegray},
  stringstyle=\color{orange},
  basicstyle=\ttfamily\tiny,
  breakatwhitespace=false
  commentstyle=\color{codegreen},
  keywordstyle=\color{codepurple},
  numberstyle=\tiny\color{codegray},
  stringstyle=\color{orange},
  basicstyle=\ttfamily\tiny,
  breakatwhitespace=false,
  breaklines=true,
  captionpos=b,
  keepspaces=true,
  numbers=left,
  numbersep=5pt,
  showspaces=false,
  showstringspaces=false,
  showtabs=false,
  tabsize=2,
  name = "Kod"
}

\title{Italian Crossword Generator: Enhancing Education through Interactive Word Puzzles}

\author{\textbf{Kamyar Zeinalipour$^{1}$, Tommaso Iaquinta$^{1}$, Asya Zanollo $^{1}$,  Giovanni Angelini$^{2}$,}\\  
\textbf{Leonardo Rigutini$^{1,2}$, Marco Maggini$^{1}$, Marco Gori$^{1}$} \\
  1. Università degli Studi di Siena, Via Roma, 56, 53100 Siena, Italy \\
  2. expert.ai, Via Virgilio, 48/H – Scala 5 41123, Modena, Italy  \\
  {\tt kamyar.zeinalipour2; marco.maggini; marco.gori@unisi.it},\\ {\tt tommaso.iaquinta; a.zanollo@student.unisi.it},\\ {\tt gangelini; lrigutini@expert.ai}}

\date{}

\begin{document}
\maketitle
\begin{abstract}
Educational crosswords offer numerous benefits for students, including increased engagement, improved understanding, critical thinking, and memory retention. Creating high-quality educational crosswords can be challenging, but recent advances in natural language processing and machine learning have made it possible to use language models to generate nice wordplays. The exploitation of cutting-edge language models like GPT3-DaVinci, GPT3-Curie, GPT3-Babbage, GPT3-Ada, and BERT-uncased has led to the development of a comprehensive system for generating and verifying crossword clues. A large dataset of clue-answer pairs was compiled to fine-tune the models in a supervised manner to generate original and challenging clues from a given keyword. On the other hand, for generating crossword clues from a given text, Zero/Few-shot learning techniques were used to extract clues from the input text, adding variety and creativity to the puzzles. We employed the fine-tuned model to generate data and labeled the acceptability of clue-answer parts with human supervision. To ensure quality, we developed a classifier by fine-tuning existing language models on the labeled dataset. Conversely, to assess the quality of clues generated from the given text using zero/few-shot learning, we employed a zero-shot learning approach to check the quality of generated clues. The results of the evaluation have been very promising, demonstrating the effectiveness of the approach in creating high-standard educational crosswords that offer students engaging and rewarding learning experiences.
\end{abstract}



\section{Introduction}

Crossword puzzles serve as a highly effective educational tool for numerous reasons. Firstly, they play a crucial role in enhancing children's vocabulary and spelling abilities, as solving the puzzles requires accurate word spelling \cite{orawiwatnakul2013crossword,dzulfikri2016application,bella2023improving}. Moreover, crossword puzzles are particularly beneficial for acquiring new lexicons in language classes and subjects that involve specialized technical terms \cite{nickerson1977crossword,sandiuc2020use,yuriev2016crossword}. Secondly, these puzzles foster problem-solving skills since students must engage in critical thinking to match clues with appropriate phrases \cite{kaynak2023effect,dol2017gpbl}. Additionally, crossword puzzles contribute to memory retention, as students need to recollect previously learned material to complete the puzzles \cite{mueller2018testing,dzulfikri2016application}. Lastly, they create an enjoyable and engaging learning experience, motivating students to continuously practice and improve their skills \cite{zirawaga2017gaming,bella2023improving}. In summary, crossword puzzles offer an enjoyable and effective approach to practice and enhance essential educational abilities \cite{zamani2021use,yuriev2016crossword}.

Creating educational crosswords requires skill, but this process can be time-consuming and limited by human resources. Recent advancements in natural language processing and machine learning offer an alternative solution: training Large Language Models (LLMs) on vast amounts of data to generate diverse and engaging crossword clues and reduce creation time.

This paper makes several contributions to the field. Our initial contribution involves the utilization of this paper to introduce an extensive dataset comprising Italian crossword clue-answer pairs, on the other hand, contributions to the field by proposing a system that uses LLMs to generate high-quality educational crossword. Our approach includes fine-tuning, zero/few-shot learning, and prompt engineering to generate clues from text and keywords. To ensure quality, we developed a set of models to filter out undesirable clues. We additionally employ an algorithm to create educational crossword schema. The resulting system can generate and filter crossword clues, creating educational crosswords with the generated clue-answer pairs.

The paper's organization is as follows: Section Two provides a comprehensive review of relevant work, and Section Three outlines the dataset used in this study. In Section Four, we detail our investigation's approach, followed by the presentation of our test findings in Section Five. Finally, Section Six concludes this study, highlighting its implications and potential future directions.

\section{Related works}

The art of crafting crossword puzzle clues has been a puzzle in itself, prompting diverse strategies to tackle the challenge. Traditional methods often lean on well-established dictionaries, thesauri, or language analysis of web-retrieved texts to define clues \cite{rigutini2008fully,rigutini2012automatic}. However, in a groundbreaking leap forward, Rigutini and colleagues unveiled the world's first fully automated crossword generator in 2008. Embracing the realm of natural language processing and machine learning, their innovative system autonomously generated crossword puzzle clues. The approach involved web crawling for documents, extracting word meanings, and utilizing techniques like part-of-speech tagging, dependency parsing, WordNet-based similarity measures, and classification models to rank clues by relevance, uniqueness, and readability.

Taking another path, \cite{ranaivo2013automatic} proposed an NLP-driven method for constructing crossword puzzles. They commenced by assembling a collection of texts related to the puzzle's theme.
Subsequently, four critical components were built: pre-processing, candidate generation, clue production, and answer selection, altogether orchestrating a comprehensive and captivating crossword puzzle.

Venturing into the realm of Spanish language puzzles, \cite{esteche2017automatic} explored extracting definitions from news articles to craft crossword puzzles. They employed a two-stage process: first, identifying crucial words and phrases and extracting their meanings from a trustworthy online dictionary, followed by utilizing those definitions as clues to construct engaging crosswords.

In another linguistic context, \cite{arora2019automatic} presented SEEKH, a software application employing natural language processing to extract keywords and craft crossword puzzles in a multitude of Indian languages. Combining statistical and linguistic tools, SEEKH adeptly pinpointed essential keywords, bringing to life a medley of crosswords across linguistic landscapes.

Despite extensive research efforts, effectively producing comprehensive and distinctive sets of clues and answers from linguistic corpora remains a formidable challenge, especially when dealing with the nuanced intricacies of the Italian language. To tackle these challenges head-on, we present an innovative methodology utilizing Language Models (LLMs) to craft sophisticated educational clues. Representing a pioneering endeavor, our approach successfully generates Italian educational crossword puzzles, addressing a void that previous methods have left unattended. By creating intellectually stimulating and original crossword puzzles, this novel technique enriches learners' profound comprehension of the subjects through detailed and encompassing answers. Therefore, our proposed work not only introduces novelty to the realm of Italian crossword generation but also provides a groundbreaking solution within the domain of educational tools.

\section{Dataset} \label{sec:dataset}
To fine-tune the LLMs, we leveraged a comprehensive collection of Italian crossword clues and answers. The sources of the clues-answer pairs are both internet sites that release solutions for crossword clues as \url{https://www.dizy.com/} and \url{https://www.cruciverba.it/} that we scraped through apposite scripts. And also \textit{pdf} versions of famous Italian crossword papers like \textit{Settimana Enigmistica} and \textit{Repubblica}, that we suitably converted to clue-answer pairs. The various sources where than cleaned, merged and the duplicates were removed. We intend to release this dataset with the support of this paper. This dataset consists of 125,600 entries that correspond to unique clue-answer pairs. It included clues related to different domains, such as history, geography, literature, and pop culture. The dataset under investigation contains a diverse array of linguistic features, including grammatical structures, syntactic patterns, and lexical elements. 

\begin{figure}
    \begin{center}
       \includegraphics[width = 0.5\textwidth, trim={0cm 0.3cm .2cm .5cm},clip]{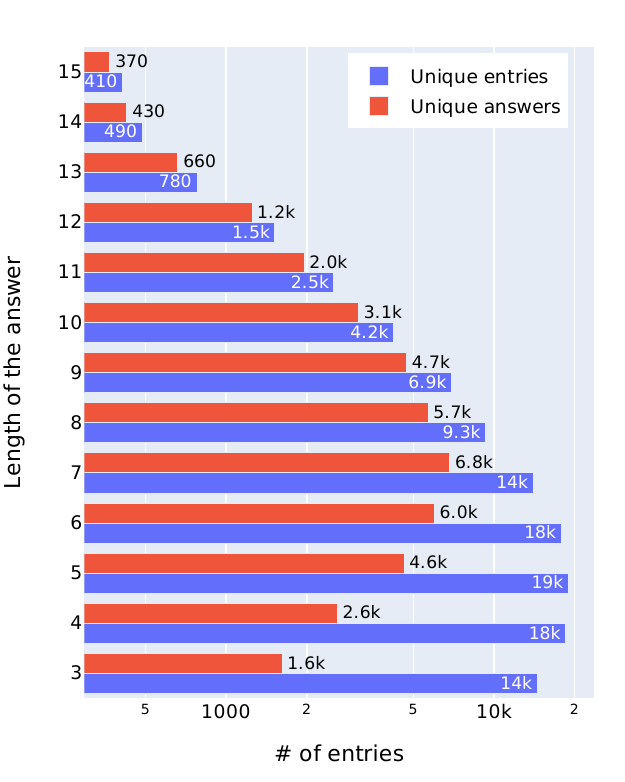} 
    \end{center}
    \caption{Distribution of the database entries by answer length, in blue the unique answer-clue pairs and in red the unique answers.} 
    \label{fig:dataset_ov}
\end{figure}

A recurring structural pattern in the dataset is the usage of the phrase ``known for" or ``used for" to define a particular place or object. For example, the definition of a certain location might be ``a place known for its historical significance" or ``an object used for a specific purpose." In both cases, the answer is a specific instance of the category described in the definition. Moreover, the dataset includes instances where the definition employs clever wordplay or exploits general category definitions to arrive at a specific answer. For example, ``In the middle of the Lake" might elicit the response ``AK", while ``An exotic legume" could be answered with ``SOY" by virtue of its membership in the broader category of legumes. In figure \ref{fig:dataset_ov} you can further go into detail regarding the distribution of the data divided by the length of the answers. Shorter answers tend to have more clues associated while as the answer gets longer the number of clues diminishes in proportion. One of the primary goals of this study was to establish the groundwork for future research by making the processed dataset publicly accessible, with the aim of encouraging other scholars to contribute to this field."\footnote{The dataset is available at \url{https://huggingface.co/datasets/Kamyar-zeinalipour/ITA_CW}}

\section{Methodology}

\begin{figure}

    \begin{center}
       \includegraphics[width = 0.5\textwidth]{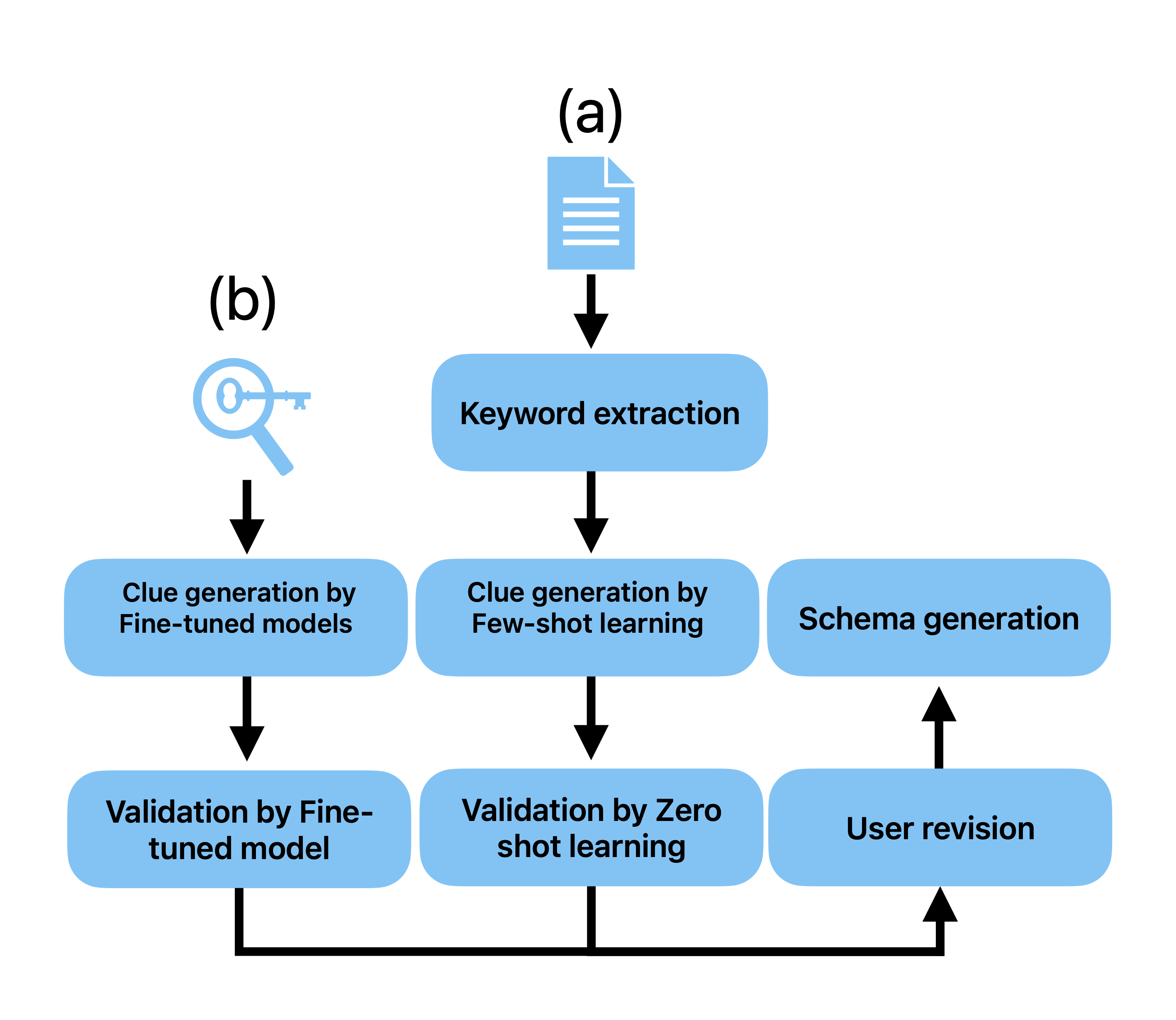} 
    \end{center}
    \caption{Overall System Architecture} 
    \label{fig:system}
\end{figure}

The system extracts clue-answer pairs from provided texts (path (a) of Figure \ref{fig:system}), or generates clues based on given keywords (path (b) of Figure \ref{fig:system}). As input texts we use paragraphs selected from Wikipedia pages on educational topics like science, geography, economics. Using this type of text allows us to create direct clues like definitions, appropriate for the educational usage. The system evaluates the quality of the generated clue-answer pairs using various validators. Following the generation process, users are granted the opportunity to review all the produced clue-answer pairs and select their preferred combinations. These selected pairs are then utilized by the final component of the system to generate the crossword puzzle schema.

In this segment, we will delve into the system's fundamental aspects, encompassing three essential components: the generation and validation of clue-answer pairs from provided text, the creation of clues based on given keywords, the validation of the result, and lastly, the generation of the crossword puzzle layout or schema.

\subsection{Path (a)}\label{sec:validation}

In this section, we analyze the path (a) of Figure \ref{fig:system}. We used a multi-step process to apply zero-shot and few-shot learning techniques to text. First, we divided the text into paragraphs and extracted precise keywords. Then, we created personalized clues inspired by the original text using those keywords. To ensure high quality, we thoroughly validated the generated clue-answer pairs. Our primary tool was the GPT-3 DaVinci base model \cite{brown2020language}. We'll explore each step in detail in the following. 

{\bf Keyword extraction}: Our innovative strategy harnesses the power of zero-shot learning for an approach to our task. We meticulously craft two prompts in both Italian and English, ensuring they are well-structured with clear objectives and detailed steps to achieve them. You can access it in the appendix under the section labeled Prompts \ref{prompt:itaprompt_kw} and \ref{prompt:engprompt_kw}. This thoughtful design empowers the Language Model (LLM) to precisely extract the most relevant keywords, capitalizing on its robust zero-shot learning capabilities. By providing guidance through our prompts, we optimize the model's ability to understand and respond to the intricacies of the task at hand

{\bf Clue generation}: We use a few-shot learning approach to create compelling crossword clues for each identified keyword in the paragraph. By leveraging an example educational text, crossword keywords, and valid clue examples, we empower the Language Model (LLM) to craft meaningful clues. We presented the paragraph and extracted clues as prompts to the LLM, allowing it to generate clues based on the provided text and keywords. This technique ensures precise and contextually relevant clues. We crafted prompts in both Italian and English, similar to the previous section. Two distinct types of prompts were developed, and all of them are accessible in the Appendix under Prompts \ref{prompt:itapromptclue} and \ref{prompt:engpromptclue}.

{\bf Validation}: We improved the quality of generated keywords and clues by implementing a multi-stage filtering process. First, we filtered out long keywords (over 3 words) as they were less suitable for crossword puzzle answers. Some generated clues inaccurately described their corresponding keywords, and some were hallucinations from the provided text. To address this, we used zero-shot learning to identify and filter out unwanted clues, resulting in a significant improvement in the final output. We created Italian and English prompts, akin to the previous section. Both prompt types can be found in the Appendix under Prompts \ref{prompt:itapromptautocheck} and \ref{prompt:engpromptautocheck}.

\subsection{Path (b)} \label{sec:models}

Referring to pipeline (b) of Figure \ref{fig:system}; addressing situations where users lack access to the original text and wish to generate crossword clues solely from given answers, we devised an approach to cater to this scenario. Our strategy encompassed multiple stages, each contributing to the overall effectiveness of the solution.

Initially, we focused on fine-tuning various language models specifically tailored for this unique task. Leveraging the data generated from these fine-tuned models, we then proceeded to create diverse classifiers. These classifiers were carefully designed with the primary objective of distinguishing high-quality clue-answer pairs from those that were deemed less suitable.

{\bf Fine-tuned models}: In the pursuit of generating crossword clues from given answers, we undertook various fine-tuning processes of language models, using data collected from Section \ref{sec:dataset}. Our selection of models comprised GPT3-DaVinci (175B parameters) and GPT3-Curie (13B parameters).

GPT3-DaVinci, with its vast parameter count, demonstrated unmatched depth, enabling it to uncover intricate patterns and craft nuanced clues. On the other hand, GPT3-Curie, while slightly smaller, proved remarkable in grasping language subtleties, further enhancing the fine-tuning process \cite{brown2020language}.

In our fine-tuning process, we employ a distinctive approach by inputting the answer and tasking the model to generate the corresponding crossword clue. This iterative method not only refines the model's ability to comprehend context but also hones its skill in crafting clues that are both challenging and contextually fitting. By continually providing the answer as input during fine-tuning, we guide the model toward a nuanced understanding of how to construct clues that align seamlessly with the given solution. This tailored training methodology further enhances the model's proficiency in delivering accurate and engaging crossword clues, solidifying its role as a versatile and effective tool in the clue-generation process.

{\bf Validation}: We developed different strong classifiers using fine-tuned language models to distinguish good crossword clues from poorly crafted ones since not all generated clues fit the given answers perfectly.

In pursuit of this goal, we fine-tuned several models, each boasting unique capacities: GPT3-DaVinci (175B parameters), GPT3-Curie (13B parameters), GPT3-Babbage (1.3B parameters), GPT3-Ada (350M parameters) \cite{brown2020language}, and BERT-uncased-base (110M parameters) \cite{raffel2020exploring}. 

By harnessing the collective power of these models, each with varying parameter counts, we gained a comprehensive perspective on their effectiveness in filtering and validating the generated clues. Through this approach, our goal was to ensure that only high-quality and contextually relevant crossword clues remained, thereby elevating the overall accuracy and usability of our system.

\subsection{Educational Crossword Schema Generator}
Our algorithm for creating educational crosswords takes input such as answer lists, work area dimensions, and stopping criteria. It starts by randomly placing a central answer, then adds other answers nearby. The algorithm iteratively adds answers, sometimes removing recent ones or restarting. The best solution is selected based on a global score of the generated schemes. Each solution produced is evaluated using the following formula:

\begin{equation*}
    \mathrm{Score} = (\mathrm{FW} + 0.5 \cdot \mathrm{LL}) \cdot \mathrm{FR} \cdot  \mathrm{LR} 
\end{equation*}

where $\mathrm{FW}$ (Filled  Words) is the number of words added; $\mathrm{LL}$ ( Linked Letters) is the number of letters that belong to two crossing words; $\mathrm{FR}$ (Filled Ratio) is the number of total letters divided by the minimum rectangle area used; and  $\mathrm{LR}$ (Linked Letters Ratio) is the  Linked Letters ($\mathrm{LL}$) divided by the number of total letters.

The algorithm incorporates various stopping criteria, including the minimum number of answers added to the grid; reaching the threshold of minimum Filled Ratio;  the limit on the number of times the grid is rebuilt from scratch, and the maximum time duration. The solution with the highest score is deemed the best. These stopping criteria play a crucial role in guiding the algorithm's decision-making process, determining when to conclude the crossword construction. Through the establishment of thresholds and limitations, we successfully ensure the efficient and effective generation of crosswords.

Within the filling process, we have the option to designate a list of "preferred answers." The algorithm places a higher priority on selecting answers from this list, increasing the probability of their incorporation into the grid. 

\section{Experiments}

The experimental evaluation of the designed system is presented in this section, focusing on the individual components and their roles in the overall framework. The system's performance is thoroughly analyzed to assess its effectiveness and efficiency, providing insights into its strengths and weaknesses.

\subsection{Experimental Evaluation: Path (a)}
In our experiments, we observed variations in model output quality when altering the language of the prompts. To demonstrate this, we conducted two sets of experiments using two types of prompts: one in English and the other in Italian. Our system underwent a rigorous evaluation process using 50 paragraphs sourced from Wikipedia to assess the performance of each component using Italian and English. Human supervision was employed, and guidelines for evaluation can be found in Appendix \ref{sec:appendix}. The results of these evaluations are summarized in Table.

Initially, our focus was on keyword extraction, and we achieved promising results in our experiments. Specifically, employing the zero-shot learning approach, we obtained 79.73\% and 75.60\% accuracy in generating suitable keywords for crossword clues using Italian and English prompts, respectively.
Subsequently, we subjected the clue-generation process to human evaluation and found that, with Italian and English prompts, 68.34\% and 76.70\% of the generated clues were considered acceptable, respectively.
To ensure the validity of our results, we employed various approaches outlined in Section \ref{sec:validation}. Through this validation, we were able to identify 56.76\% and 69.72\% of the unacceptable clue-answer pairs generated using the Italian and English prompts, respectively. These results clearly demonstrate the effectiveness of our system in producing satisfactory crossword clues based on the evaluated text.

\begin{table*}[htbp]
\caption{Assessment outcomes of the clue-answer pairs generated from the provided Text.}

\begin{center}
\begin{tabular}{ccc}
\toprule
\textit{System part } & \textit{Italian Prompt} & \textit{English Prompt} \\
\midrule
Acceptable keywords &   79.73 \%    &   75.60\% \\
Acceptable clues    &   68.34 \%    &   76.70 \% \\
Validator performance   &   56.76 \%    &   69.72 \% \\
\bottomrule
\end{tabular}
\label{tab:prompt}
\end{center}
\end{table*}

\begin{figure*}[h]
    \centering
    \includegraphics[width=\textwidth]{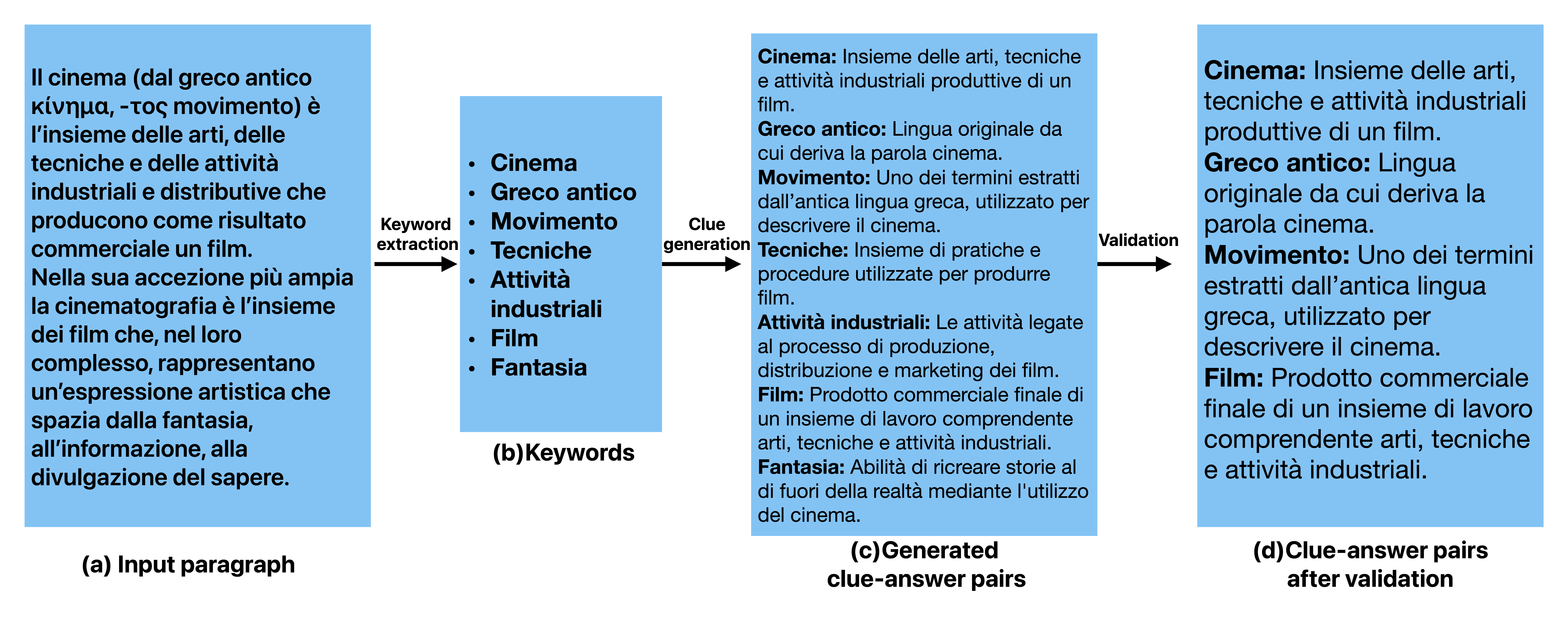}
    \caption{An concrete example of the path (a)}
    \label{fig:pipelineexample}
\end{figure*}

Figure \ref{fig:pipelineexample} demonstrates the step-by-step process of generating crossword clue-answer pairs from input text. The image shows the various stages, such as keyword extraction, clue creation, and pair validation, and illustrates how our system converts input text into pertinent crossword clues.
The results with the Italian data revealed that, when the prompt is in English, the performance of the model is better than when the prompt is in Italian.

\subsection{Experimental Evaluation: Path (b)}

This section delves into our experimental endeavors on generating and validating clues from keywords. Building upon the insights presented in Section \ref{sec:models}, we devised and fine-tuned two distinct models GPT3-DaVinci and GPT3-Curie with a specific focus on creating clues based on given keywords. For the training phase, we selected a subset of the dataset introduced in Section \ref{sec:dataset}, encompassing 50000 unique clue-answer pairs.

Once the fine-tuning phase concluded, we generated 4,000 clues from each of the fine-tuned models and subjected them to human evaluation using the guidelines provided in Appendix \ref{sec:appendix}. The outcomes of this evaluation are summarized in Table \ref{tab:pathb}. Remarkably, GPT-3 DaVinci outperformed GPT-3 Curie, yielding an impressive 60.1\% of acceptable clues compared to Curie's 34.9\%

\begin{table}[htbp]
\caption{Assessment outcomes of the clues generated from the provided keyword.}
    \begin{center}
        \begin{tabular}{cc}
        \toprule
        \textit{Model}&  \textit{\% of acceptable  clues}  \\
        \midrule
        GPT3-DaVinci& 60.1   \\
        GPT3-Curie& 34.9  \\
        \bottomrule
        \end{tabular}
        \label{tab:pathb}
    \end{center}
\end{table}

To gain deeper insights into the quality of the generated clues, we meticulously assembled a collection of acceptable and unacceptable clues. These were randomly sampled from the human-supervised label dataset, offering a diverse clue for each answer. Please consult Table \ref{tab:valexample} (refer to table \ref{tab:tabtranslate} in the Appendix for translation). This detailed analysis helps us evaluate the quality and suitability of the clues for creating engaging crossword puzzles.

\begin{table*}[h]
\caption{Acceptable and unacceptable clues from given keywords using various models.}

\begin{center}
\begin{tabular}{ccc}
\toprule
\textit{Clue-Answer pair} & \textit{Model} & \textit{Accepted}  \\
\midrule
Mitologia: La conosce chi conosce i miti & DaVinci & Yes \\
Elettricità: Uno dei segni zodiacali & DaVinci & No \\
Curiosità: Il desiderio di sapere & Curie & Yes \\
Collaborazione: Lo si raggiunge con chiunque & Curie & No \\

\bottomrule
\end{tabular}
\label{tab:valexample}
\end{center}
\end{table*}

We developed multiple classifiers that integrate different language models to differentiate between acceptable and unacceptable clue-answer pairs. The result of the analysis on the test set is shown in Table \ref{tab3}.
We utilized a dataset of 6,000 human evaluations from the previous step to construct various classifiers. This is the data which we tried to evaluate GPT-3-Davinci and GPT-3-Curie by human supervision. For training and evaluation, we employed 80\% of this data for fine-tuning the classifiers and reserving the remaining 20\% for testing the classifiers. Within the dataset, 51\% comprised acceptable clues, while the remaining 49\% consisted of unacceptable clues.

\begin{table*}[htbp]
\caption{Classifier performance on distinguishing acceptable Clue-Answer pairs}

\begin{center}
\begin{tabular}{ccccc}

\toprule
\textit{Model}&  \textit{accuracy \%} & \textit{precision \%}& \textit{recall \%}& \textit{F1 Score}  \\
\midrule
GPT3-Dvinci & 79.88  & 80.16    & 76.67  & 0.7838\\
GPT3-Curie  & 77.82  & 78.80    & 72.99  & 0.7578\\
GPT3-Babbage& 74.12  & 72.58    & 73.25  & 0.7291\\
GPT3-Ada    & 69.17  & 67.77    & 67.06  & 0.6741\\
BERT-uncased-base       & 65.62  & 63.71    & 64.47  & 0.6409\\
\bottomrule
\end{tabular}
\label{tab3}
\end{center}
\end{table*}

The evaluation results reveal significant distinctions among the classifiers in their ability to differentiate between acceptable and unacceptable clue-answer pairs. Earning the top position, the GPT3-DaVinci model achieved an accuracy of 79.88\%, solidifying its role as the most effective classifier in this task. Following closely, the GPT3-Curie base model attained a commendable 77.82\% accuracy. The GPT3-Babbage model demonstrated respectable performance with 74.12\% accuracy, while GPT3-Ada and BERT-uncased achieved accuracies of 69.17\% and 65.62\%, respectively.

\subsection{Schema Generation}
\begin{figure}[hht]
    \begin{center}
       \includegraphics[width = 0.5\textwidth]{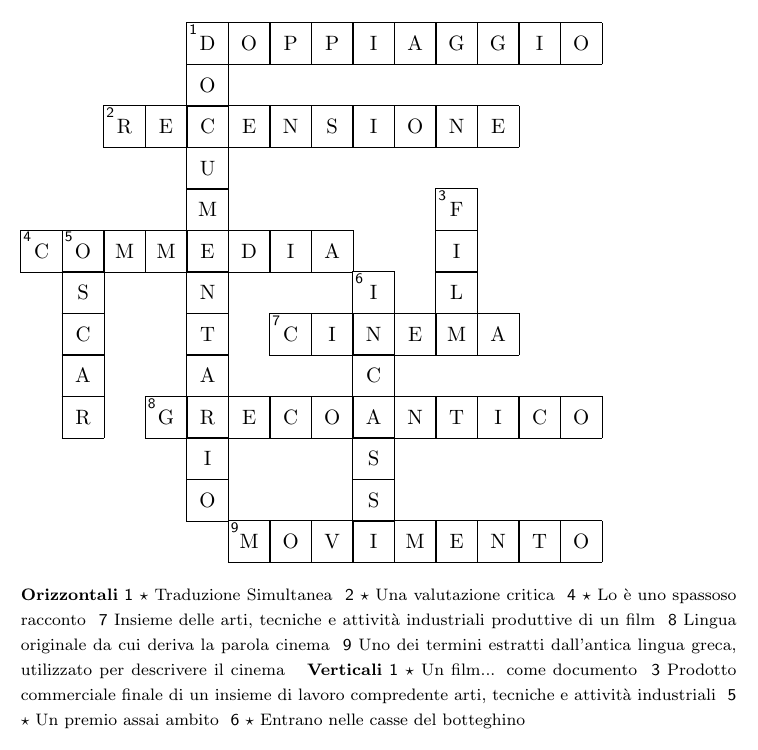} 
    \end{center}
    \caption{An illustrative crossword created using the newly introduced system.  }
    \label{fig:crossword}
\end{figure}

Our schema generation algorithm creates educational crosswords with diverse layouts using a single batch of words. Below is an illustration, check the Figure \ref{fig:crossword} of a comprehensive Italian educational crossword about movies produced with our system. The clue-answer pairs are both extracted from a text (path (a), see Figure \ref{fig:pipelineexample}) and generated directly from a keyword (path (b), contr-assigned with a $\star$ below). 

\section{Conclusions}
In this paper, we present various contributions, including the introduction of a substantial dataset for Italian clue-answer pairs, we developed an innovative system using Large Language Models to generate educational crossword puzzles from given texts or answers. Our approach combines human supervision and specific guidelines to ensure high-quality and relevant clues. 

Our system includes a keyword extraction component (79.73\% high-quality keywords) and a crossword clue generation component (76.6\% relevant and acceptable clues). A validation component filters out unacceptable pairs, achieving a 69.72\% detection rate. We conducted an in-depth investigation of fine-tuned generators and classifiers to enhance the quality of clues. Among the models tested, GPT3-Davinci demonstrated exceptional performance in generating clues based on given keywords, producing a remarkable 60.1\% of acceptable clues. Moreover, GPT3-Davinci proved to be the most proficient classifier, accurately distinguishing between good clue-answer pairs and unacceptable ones with an impressive 79.88\% accuracy.

Our algorithm for generating educational crossword schemes is efficient and produces diverse layouts. This study aims to enhance student skills and promote interactive learning. Educators can integrate our system into their instruction for more effective teaching practices.

Future research involves developing advanced models for direct clue-answer pair generation and exploring specialized models for different clue types. Our vision is to revolutionize educational crossword generation and unlock new innovations in teaching practice.

\section*{Acknowledgments}
This work was supported by the IBRIDAI project, a project financed by the Regional Operational Program "FESR 2014-2020" of Emilia Romagna (Italy), resolution of the Regional Council n. 863/2021.

\bibliographystyle{acl}
\bibliography{bibliography.bib}

\section*{Appendix}\label{sec:appendix}

\subsection*{Guidelines for Validating Clue-Answer Pairs}
In the course of our study, we embraced an enthralling challenge: constructing a classifier capable of discerning between acceptable and non-acceptable crossword clue-answer pairs. Crossword puzzles have held a cherished place as a beloved pastime, demanding a harmonious fusion of linguistic prowess, creative acumen, and adherence to intricate puzzle construction rules to fashion top-tier clue-answer pairs. Our pursuit of creating an automatic evaluator for generated crossword clues and their corresponding answers holds tremendous potential. This advancement promises to aid puzzle creators, enrich puzzle-solving experiences, and unlock profound insights into the subtle nuances of language and puzzle design. Ultimately, this endeavor not only elevates the world of crossword puzzles but also kindles a deeper appreciation for their linguistic artistry and cognitive allure.

To create a powerful classifier for crossword clue-answer pairs, we must establish a strong and comprehensive guideline that clearly delineates the attributes of acceptable and non-acceptable pairs. This guideline will be the cornerstone for training our classifier, enabling it to discern the defining characteristics that set apart high-quality clues from irrelevant or inappropriate ones. With strict adherence to this guideline, we can guarantee the accuracy of our classifier in assessing the quality of clue-answer pairs, ultimately leading to the creation of more captivating and enjoyable crossword puzzles.

Let us now explore the pivotal components of the guideline, essential for evaluating crossword clue-answer pairs:

\textbf{Relevance and Cohesion}: A top-notch crossword clue-answer pair thrives on a profound and meaningful connection between the clue and the answer. The clue should provide ample context or clever hints that smoothly lead solvers to the intended solution. Simultaneously, the answer must be directly tied to the clue, fitting flawlessly within the puzzle's theme or topic.

\textbf{Wordplay and Inventiveness}: Elevate your crossword clues with ingenuity and wordplay that challenge and delight solvers. Seek clues that encourage lateral thinking, incorporate witty twists, or conceal intriguing meanings. A well-crafted clue-answer pair captures the solver's imagination, transforming the puzzle into an exhilarating journey of discovery.

\textbf{Clarity and Precision}: Precision is key in creating crossword clues. Ensure your clues are crystal clear and unambiguous, presenting solvers with a distinct and precise solution. Avoid any ambiguity that might lead to multiple interpretations or numerous possible answers. The goal is to deliver a single correct solution that aligns perfectly with the clue's intended meaning.

\textbf{Grammar and Language}: Pay meticulous attention to grammar, syntax, and linguistic conventions in both the clue and the answer. Maintain grammatical correctness, coherence, and an appropriate level of complexity for a crossword puzzle.

\textbf{General Knowledge and Fairness}: Strike a balance between challenge and accessibility by grounding your clues in general knowledge or commonly known facts. Avoid overly obscure or specialized references that could alienate solvers. A great clue-answer pair caters to a diverse range of puzzle enthusiasts, offering a fair and engaging experience for all.

Through the adoption of this framework, a robust dataset can be generated, facilitating the development of a dependable classifier that discerns commendable crossword clue-answer pairs from incongruous or inappropriate ones. This transformative classifier holds the promise of revolutionizing crossword puzzle creation, assessment, and solving, offering invaluable revelations into the craft of constructing captivating and mentally stimulating puzzles.

\begin{table*}[h]
\caption{Translation of Table \ref{tab:valexample}}

\begin{center}
\begin{tabular}{ccl}
\hline
\textit{Clue-Answer pair} & \textit{Model} & \textit{Acc.}  \\
\hline
Mythology: It is known by anyone who knows myths & DV & Yes \\
Electricity: One of the zodiac signs & DV & No \\
Curiosity: The desire to know & Curie & Yes \\
Collaboration: One reaches it with anyone & Curie & No \\
\hline
\end{tabular}
\label{tab:tabtranslate}
\end{center}
\end{table*}

\subsection*{Prompts}
\begin{lstlisting}[caption={\textbf{Italian, for keyword extraction}}, captionpos=t, label=prompt:itaprompt_kw]
    prompt = f"""
    Obiettivo: Il tuo compito è estrarre delle parole chiave, descritte nel testo proposto. Le parole chiave estratte saranno utilizzate per creare brevi definizioni di cruciverba riguardanti il testo da cui sono estratte le parole chiave. Le definizioni saranno d'aiuto per trovare la soluzione corrispondente e completare il cruciverba.

    Completa l'obiettivo attraverso i seguenti passaggi:

    1- Estrai le parole chiave più importanti del testo.

    2- Controlla le parole chiave: controlla se le parole chiave sono descritte e definite nel testo o non sono descritte e definite nel testo.

    3- Parole chiave finali : sulla base del passaggio precedente, rimuovi tutte le parole chiave che non sono definite nel testo.



    Utilizza il seguente formato di output:

    Parole chiave: <Parole chiave finali>


    Text: ```{text}```
    """
\end{lstlisting}

\begin{lstlisting}[caption={\textbf{Italian, for clue generation}}, captionpos=t, label=prompt:itapromptclue]
     prompt = f"""
    Genera brevi definizioni di cruciverba per ciascuna delle parole chiave fornite: {keywords} sulla base del seguente testo: {text}.

    Completa l'obiettivo attraverso i seguenti passaggi:

    1- Per ciascuna delle parole chiave fornite, trova il passaggio del testo contentente l'informazione riguardante la parola chiave.
    2- Genera brevi definizioni: per tutte le parole chiave genera brevi definizioni riguardanti il testo. Nella definizione non deve essere presente la parola chiave.
    3- Non usare virgolette e apostrofi nell'output.

    Segui questo esempio per completare l'obiettivo: 
    "Testo: La scienza è un sistema di conoscenze ottenute attraverso unattività di ricerca prevalentemente organizzata con procedimenti metodici e rigorosi, coniugando la sperimentazione con ragionamenti logici condotti a partire da un insieme di assiomi, tipici delle discipline formali. Uno dei primi esempi del loro utilizzo lo si può trovare negli Elementi di Euclide, mentre il metodo sperimentale, tipico della scienza moderna, venne introdotto da Galileo Galilei, e prevede di controllare continuamente che le osservazioni sperimentali siano coerenti con le ipotesi e i ragionamenti svolti.
    Parole chiave: conoscenze, ricerca, rigorosi, assiomi, ipotesi, Galileo
    Definizioni: 
    Conoscenze: informazioni acquisite tramite ricerca organizzata con procedimenti metodici e rigorosi.
    Ricerca: attività organizzata prevalentemente con procedimenti metodici e rigorosi finalizzata allottenimento di conoscenze.
    Rigorosi: esatti e precisi nello svolgimento delle azioni. 
    Assiomi: un insieme di verità accettate come base dei ragionamenti logici.
    Ipotesi: assunte per comprendere le osservazioni sperimentali e testare le conoscenze
    Galileo : egli introdusse il metodo sperimentale nel processo di scienza moderna.
    "
    
    
    """
\end{lstlisting}

\begin{lstlisting}[caption={\textbf{Italian, to auto check}}, captionpos=t, label=prompt:itapromptautocheck]
prompt = f"""


    Obiettivo: il tuo obiettivo è controllare se il contenuto di ogni definizione è presente o no nel testo proposto Per ciascuna definizione scrivi "True" se il contenuto è presente nel testo e "False" se il contenuto non è contenuto nel testo.


    Sentences: ```{clue}```

    Text: ```{text}```
    """
\end{lstlisting}

\begin{lstlisting}[caption={\textbf{English, for keyword extraction}}, captionpos=t, label=prompt:engprompt_kw]
 prompt = f"""


    Objective: Your task is to extract described keywords in Italian from a given Italian text. These keywords will be used to create Italian crossword short definitions based on the extracted text. The clues will help Italian solvers to find the corresponding answers and complete the puzzle grid.

    Please follow these steps to achieve the objective:

    1- Extract the most important Italian keywords in the Italian text.

    2- Check keywords: check if the Italian keywords are well Explained in the given Italian text or not.

    3- Final keywords : Remove all the Italian keywords which are not well defined in the Italian text based on the last step.


    Use the following output format:

    Keywords: <Final keywords>


    Text: ```{text}```
    """
\end{lstlisting}

\begin{lstlisting}[caption={\textbf{English, for clue generation}}, captionpos=t, label=prompt:engpromptclue]
prompt = f"""


    Generate short crossword definitions in Italian for each provided Italian keyword: {keywords} based on the following Italian text: {text}. 

    Follow these steps to achieve the objective:

    1- For each provided Italian keyword detect the part of the Italian text which contains the keyword information. 
    2- Generate short definitions in Italian: For all the Italian keywords generate short definitions in Italian based on the Italian text, and place the correspondent keyword after each generated definition. Make sure that the Italian keyword is not present in the correspondent definition.
    3- Do not use quotation marks and apostrophes in the output.

    Follow this example to complete the task: 
    "Text: La scienza è un sistema di conoscenze ottenute attraverso unattività di ricerca prevalentemente organizzata con procedimenti metodici e rigorosi, coniugando la sperimentazione con ragionamenti logici condotti a partire da un insieme di assiomi, tipici delle discipline formali. Uno dei primi esempi del loro utilizzo lo si può trovare negli Elementi di Euclide, mentre il metodo sperimentale, tipico della scienza moderna, venne introdotto da Galileo Galilei, e prevede di controllare continuamente che le osservazioni sperimentali siano coerenti con le ipotesi e i ragionamenti svolti.
    Keywords: conoscenze, ricerca, rigorosi, assiomi, ipotesi, Galileo
    Clues: 
    Conoscenze: informazioni acquisite tramite ricerca organizzata con procedimenti metodici e rigorosi.
    Ricerca: attività organizzata prevalentemente con procedimenti metodici e rigorosi finalizzata allottenimento di conoscenze.
    Rigorosi: esatti e precisi nello svolgimento delle azioni. 
    Assiomi: un insieme di verità accettate come base dei ragionamenti logici.
    Ipotesi: assunte per comprendere le osservazioni sperimentali e testare le conoscenze
    Galileo : egli introdusse il metodo sperimentale nel processo di scienza moderna.
    "

    """
\end{lstlisting}

\begin{lstlisting}[caption={\textbf{English, to auto check}}, captionpos=t, label=prompt:engpromptautocheck]
prompt = f"""


    Objective: Your objective is to check whether each given Italian Sentence content is present in the provided Italian text or not. Print "True" if it is present in the provided Italian text and "False" if it is not present in the provided Italian text.

    Sentences: ```{clue}```

    Text: ```{text}```
    """
\end{lstlisting}

\end{document}